\def\year{2022}\relax
\documentclass[letterpaper]{article} % DO NOT CHANGE THIS
\usepackage{flushend}
\usepackage{aaai22}  % DO NOT CHANGE THIS
\usepackage{times}  % DO NOT CHANGE THIS
\usepackage{helvet}  % DO NOT CHANGE THIS
\usepackage{courier}  % DO NOT CHANGE THIS
\usepackage[hyphens]{url}  % DO NOT CHANGE THIS
\usepackage{graphicx} % DO NOT CHANGE THIS
\usepackage{natbib}  % DO NOT CHANGE THIS AND DO NOT ADD ANY OPTIONS TO IT
\usepackage{caption} % DO NOT CHANGE THIS AND DO NOT ADD ANY OPTIONS TO IT
\DeclareCaptionStyle{ruled}{labelfont=normalfont,labelsep=colon,strut=off} % DO NOT CHANGE THIS
\usepackage[switch]{lineno}
\usepackage{flushend}
\usepackage{algorithm}
\usepackage{algorithmic}
\usepackage{graphicx}
\usepackage{amsmath}
\usepackage{amssymb}
\usepackage{subfigure}
\usepackage{bbding}
\usepackage{bbm}
\usepackage{newfloat}
\usepackage{listings}
\floatstyle{ruled}
\newfloat{listing}{tb}{lst}{}
\floatname{listing}{Listing}

\usepackage{expl3}
\ExplSyntaxOn
\newcommand\latinabbrev[1]{
	\peek_meaning:NTF . {% Same as \@ifnextchar
		#1\@}%
	{ \peek_catcode:NTF a {% Check whether next char has same catcode as \'a, i.e., is a letter
			#1.\@ }%
		{#1.\@}}} 
\ExplSyntaxOff

\def\ie{\latinabbrev{i.e}}

\title{Hybrid Instance-aware Temporal Fusion for Online Video Instance Segmentation} 
\author{
    Xiang Li, \textsuperscript{\rm 1,2}\thanks{This work was done when Xiang Li was an intern at MSRA.}
    Jinglu Wang, \textsuperscript{\rm 2}
    Xiao Li,\textsuperscript{\rm 2}
    Yan Lu \textsuperscript{\rm 2}
}

\affiliations{
    \textsuperscript{\rm 1} Department of Electrical and Computer Engineering, Carnegie Mellon University\\
    \textsuperscript{\rm 2} Microsoft Research Asia\\
    xl6@andrew.cmu.edu, \{jinglwa, xili11, yanlu\}@microsoft.com
}

\usepackage{bibentry}
\begin{document}
\maketitle

\begin{abstract}
Recently, transformer-based image segmentation methods have achieved notable success against previous solutions. While for video domains, how to effectively model temporal context with the attention of object instances across frames remains an open problem. In this paper, we propose an online video instance segmentation framework with a novel instance-aware temporal fusion method. We first leverage the representation \cite{wang2021max}, \ie, a latent code in the global context (instance code) and CNN feature maps to represent instance- and pixel-level features. Based on this representation, we introduce a cropping-free temporal fusion approach to model the temporal consistency between video frames. Specifically, we encode global instance-specific information in the instance code and build up inter-frame contextual fusion with hybrid attentions between the instance codes and CNN feature maps. Inter-frame consistency between the instance codes is further enforced with order constraints. By leveraging the learned hybrid temporal consistency, we are able to directly retrieve and maintain instance identities across frames, eliminating the complicated frame-wise instance matching in prior methods. Extensive experiments have been conducted on popular VIS datasets, i.e. Youtube-VIS-19/21. Our model achieves the best performance among all online VIS methods. Notably, our model also eclipses all offline methods when using the ResNet-50 backbone.

\end{abstract}

\section{Introduction}

Video instance segmentation (VIS), aiming at simultaneously classifying, segmenting, and tracking object instances, attracts increasing attention recently due to boosting interest in video analysis. VIS methods are categorized into offline and online methods according to two kinds of inputs, \ie, clip- and frame-wise inputs respectively.
Offline methods obtain impressive accuracy thanks to modeling spatial-temporal correlation throughout the whole clip \cite{bertasius2020classifying, athar2020stem, wang2021end}, but they inevitably show limitations on real streaming applications.
Online methods are more practical for streaming applications, but their performance of existing methods are far from that of offline methods because of their imperfection in modeling frame-to-frame (F2F) communications. We focus on F2F communications for online VIS task.

Some online methods \cite{yang2021crossover,fu2020compfeat,li2021video} model F2F communication only at pixel level. Modeling higher level semantics including instance-aware information is limited and expensive. 
Other methods \cite{li2021spatial,gong2021temporal,fu2020compfeat} model F2F communication at instance level by first cropping out RoI features with detected boxes to obtain instance-level proxies and then associating or fusing such cropped features. The F2F communication of box-based method heavily relies on the detection accuracy. Although sophisticated feature alignment can be employed, the F2F communication is still incomplete and biased because the cropped instances are isolated from the global context.
% To our best knowledge, there are no methods modeling the F2F communication at both pixel- and instance-level while keeping global contexts.
All prior methods either model the F2F communication at solely pixel or instance level, while no joint communications at hybrid levels are discussed.

\begin{figure}[t]
\centering
\includegraphics[width=\linewidth]{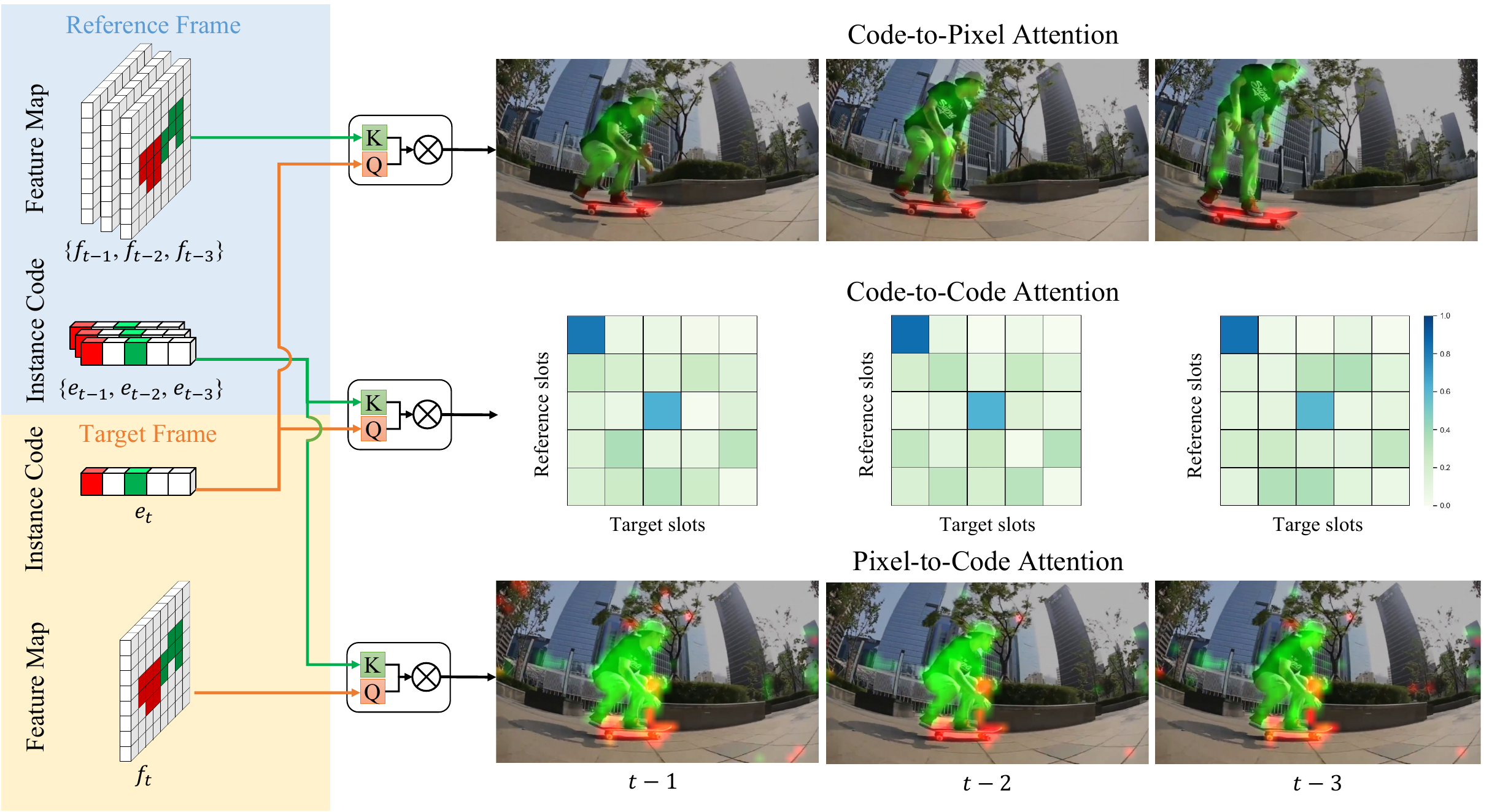}
\caption{Visualization of attention maps in \textbf{inter-frame attention} layers. The red and green colors in the instance code and feature map indicate different instances. ``Q'' and ``K'' represent the Query and Key operations respectively. All queries come from the target frame and keys from reference frames. We can find the attention responding on references $\{t-1, t-2, t-3\}$ are consistent.
}
\label{fig:teaser}
\vspace{-0.2cm}
\end{figure}

In this work, we focus on improving online VIS by introducing novel hybrid instance-aware F2F communications.
Considering the limitation of box-based methods in video frame communication, we build up our method based on the state-of-the-art box-free image segmentation framework, \ie, MaX-DeepLab~\cite{wang2021max}.
%, which by leveraging and extending the powerful instance representation from MaX-Deeplab~\cite{wang2021max} applied in box-free image segmentation. 
Inspired by Max-DeepLab, we employ a global latent code as well as CNN feature maps for jointly representing instance-aware features. We term the global code \textit{instance code} as it could capture instance-aware high-level clues in the global context in our VIS task. 
Benefiting from this hybrid representation, we enforce the temporal consistency in both instance code and feature map by two designs. 
First, we model the F2F communications at hybrid level by employing cross-attentions between both instance code and feature maps. Second, we enhance the cross-frame consistency of instance features by utilizing and further consolidating the slot consistency of instance code during training with a novel consistency constraint. In inference stage, the instance identity can be direct associated with slot indices, thus greatly reduce the cross-frame instance matching cost.

Our method is designed with two key insights.
First, adopting such a box-free representation to VIS task enables us to remove the reliance on “detection-and-cropping” approach for instance-level F2F communication; instead, we are able to build up a more comprehensive and expressive inter-frame communication at hybrid pixel and instance level with unified attention operations.
Figure~\ref{fig:teaser} illustrates an example of the inter-frame attentions between instance code and feature maps. The instance code activates pixels belonging to the same instance across frames; instance code tends to activate other codes of the same instance; pixels of the same instance get feedback attention from instance code. 
Second, recent literature \cite{carion2020end,wang2021max} has shown that the instance code tends to be ordered and correlated with each other regarding to instance position and class. This inherent property de facto fits the instance consistency for VIS task in the sense that instance prediction tend to be consistent across frames.

%Second, recent literature has shown that box-free image segmentation methods~\cite{carion2020end,wang2021max} already shown some extents of instance identities consistency across frames with per-frame inference. Interestingly, the instance code used in these methods are also tends to be ordered and correlated with each other (i.e., the code slots tend to be consistent across frames). 
%Such observations implies us that the instance code slot consistency highly contributes to the final instance identity consistency, and thus should be explicitly constrained during training.

Our main contributions are summarized as follows:
\begin{itemize}
    \item We propose a box-free, detection-free and matching-free online VIS framework. It abandons complicated matching operations as well as heavy decoders for matching instance identities in previous online VIS methods.
    \item We introduce an instance-aware temporal fusion method, which enables instance-level feature aggregation without cropping feature maps, bringing a new way to combining historical information for instance segmentation.
    \item Our model achieves state-of-the-art results on Youtube-VIS 2019 and Youtube-VIS 2021 with 41.3 mAP and 35.8 mAP respectively, outperforming previous state-of-art methods by a large margin.
\end{itemize}

%Another key insight is that code slots tend to be learned ordered and correlated with each other \cite{carion2020end,wang2021max}, indicating that such slots tend to be consistent across frames. This inherent property de facto fits the instance consistency in videos, \ie, the same instance in adjacent frames has close positions, similar appearances, and same class labels. We model F2F communication by utilizing and further consolidating this code consistency, and thus additional matching cost for instances cross frames is reduced.
%and this allows us to discard handcrafted matching rules \cite{yang2019video} and save additional matching cost.
% The advantages of this representation are
% 1) modeling joint pixel-level and instance-level communication without cropping
% 2) heads are implicitly to be learned ordered and correlated with each other.

\section{Related Works}
\subsection{Video Instance Segmentation}
Video instance segmentation requires classifying and segmenting each instance in a frame and assigning the same instance with the same identity across frames. The methods proposed for the VIS task work either in an online or offline fashion. For online methods, Mask-Track-RCNN \cite{yang2019video} is the first attempt to address the VIS problem which extends the Mask-RCNN \cite{he2017mask} with a tracking head to associate instance identities. Followed by Mask-Track-RCNN, SipMask \cite{cao2020sipmask} and SG-Net \cite{liu2021sg} build the tracking head on top of the modified one-stage still-image instance segmentation method FCOS \cite{tian2019fcos} and Blender-Mask \cite{chen2020blendmask} to achieve better speed and performance. CrossVIS \cite{yang2021crossover} introduces the cross-over learning scheme and global instance embedding to learn better features for robust instance tracking and segmentation. Different from online methods, another track of VIS takes the entire video as input and works in an offline fashion. MaskProp leverages Hybrid Task Cascade Network \cite{bertasius2020classifying} and heavily post-processing to associate and refine the predictions from Mask-RCNN \cite{he2017mask} for VIS task. More recently, VISTR \cite{wang2021end} brings a new way to tackle the VIS problem, which utilizes a transformer encoder on top of the convolutional backbone and match instance identities by transformer decoder. In VISTR, the same instances are predicted by a set of pre-defined slots in the instance queries thus it can inference in an end-to-end manner without redundant matching operations.

% Moreover, TROI \cite{gong2021temporal}, compFeat \cite{fu2020compfeat} and STMask \cite{li2021spatial} focus on temporal feature aggregation and proposes several temporal ROI feature calibration approaches to fuse pixel feature map in an instance level.

\subsection{Vision Transformer}
Transformer \cite{vaswani2017attention}, proposed for neural machine translation, has been widely used in the computer vision field. The transformer shows great generalization and has been adapted in multiple tasks, such as classification \cite{chen20182,bello2019attention,ramachandran2019stand}, image segmentation \cite{wang2021max,wang2021end}, object detection \cite{carion2020end,zhu2020deformable}, image generation \cite{parmar2018image, cornia2020meshed,yang2020learning} and video recognition \cite{neimark2021video,girdhar2019video}. Since the attention mechanism and fully connection layers (FC) in transformer make it computational expensive, most methods only apply transformer on down-sampled feature maps. For example, DETR \cite{carion2020end} and VISTR \cite{wang2021end} overlap a transformer on top of a CNN to conduct end-to-end object detection and segmentation respectively. On other hand, MaX-Deeplab \cite{wang2021max} inserts the transformer into the CNN and enables communication between different stages of it.

\section{Method}
\begin{figure*}[t]
\centering
\includegraphics[width=\textwidth]{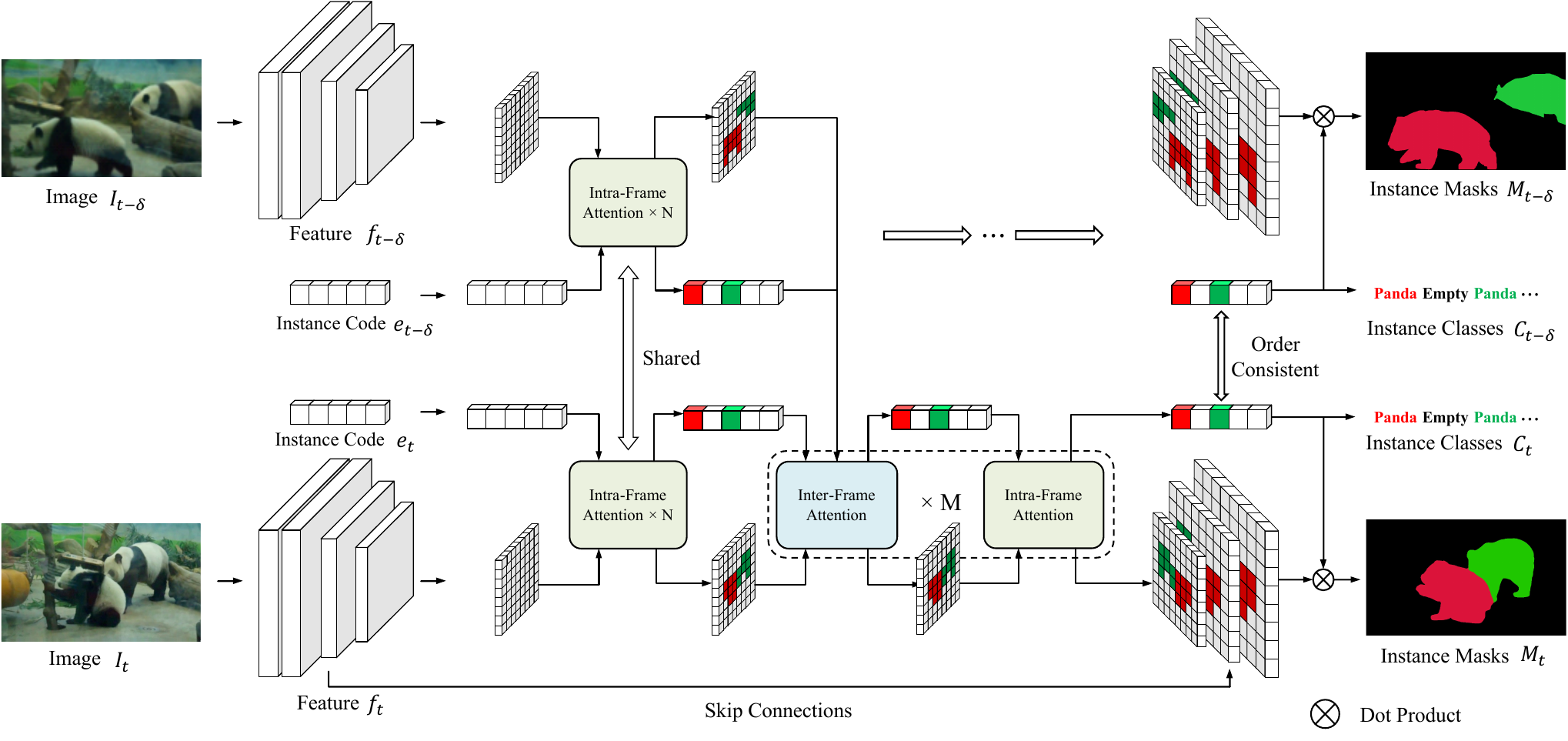}
\caption{\textbf{Overview of the proposed framework.} 
We enforce the temporal consistency in VIS by introducing hybrid F2F communications.
Two main components are highlighted, \ie, intra-frame attention for linking current instance code and feature maps, and inter-frame attention for fusing hybrid (pixel- and instance-level) temporal information in adjacent frames. The first $N$ intra-frame attention layers are integrated into the convolutional backbone followed by $M$ alternate attention layers. The final instance codes are constrained to be consistent across frames.
% The final instance masks are discriminated by dynamic filters learned from instance code. 
% To best visualize, we omit the convolution and fully connected layers in the figure.
}
\label{fig:overview}
\vspace{-0.4cm}
\end{figure*}
\subsection{Overview}
We leverage the transformer-based network to tackle the VIS problem in a bottom-up online fashion. 
Unlike previous methods~\cite{wang2021end} which simply stack a transformer on top of the convolutional neural network (CNN), we insert the transformer layers in the CNN by leveraging a stand-alone latent code to encode the instance information.
Figure~\ref{fig:overview} shows an overview of our method.
We utilize intra-frame attention layers to extract instance code for the current frame and propagate the historical information by involving inter-frame and intra-frame attention layers alternatively. 
We then use skip connections~\cite{ronneberger2015u} to get low-level contextual information and use dynamic convolution~\cite{jia2016dynamic, tian2020conditional} to generate the final segmentation maps. Here, no sophisticated matching or post-processing is required for instance identity matching as they directly correspond to the (frame-consistent) code slot indices.

%To obtain the final instance segmentation, we use skip connections \cite{ronneberger2015u} to get low-level contextual information and dynamic convolution \cite{jia2016dynamic, tian2020conditional} to distinguish the pixels to their corresponding instance. The instance identity can be easily obtained by its corresponding code slots index. Thus, no sophisticated matching or post-processing is needed.

\begin{figure}[h!]
\centering
\includegraphics[width=\linewidth]{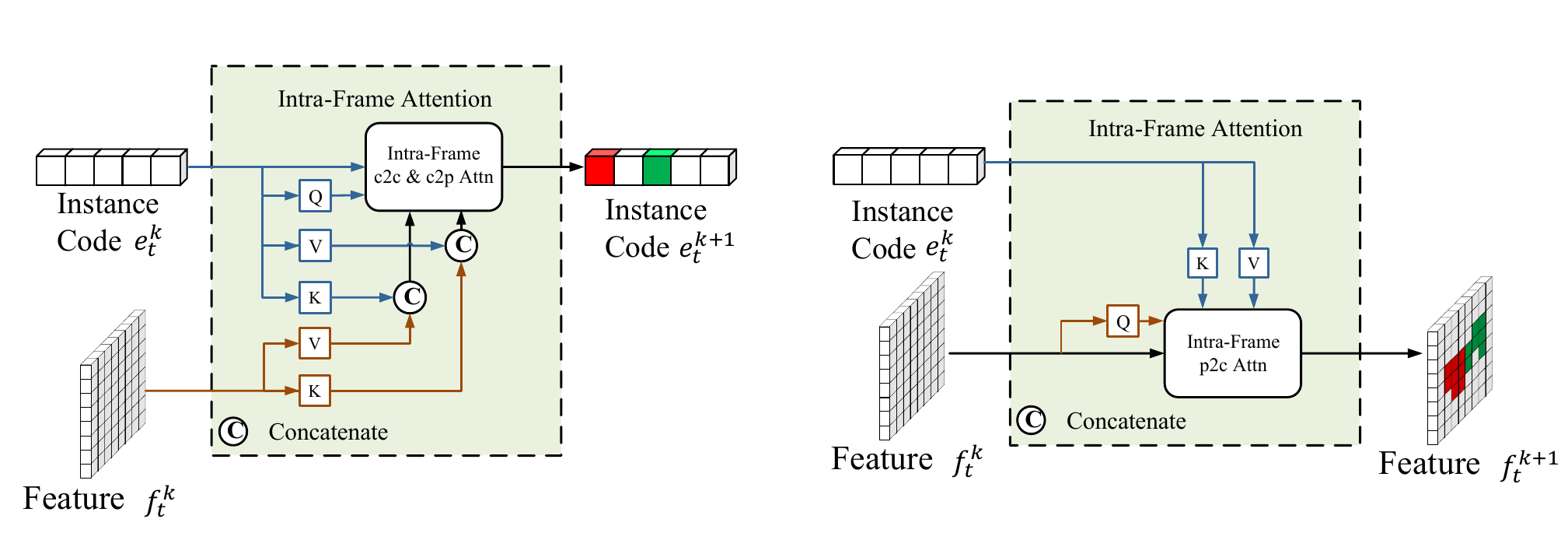}
\caption{\textbf{Intra-frame attention.} The intra-frame attention at  $k$-th layer links current frame instance code $e_t^k$ and features $f_t^k$. The c2c\,\&\,c2p attention probes high-level instance relevant features and fuses them back to the instance code. The p2c attention adjusts the pixel feature map based on the instance code.
}
\label{fig:intra_attn}
\end{figure}

\subsection{Hybrid Representation for Video Frame}
We unitize a global latent code (i.e. instance code) as well as CNN feature maps to jointly represent instance-aware features for each frame.
%, and further model correlation of frames using such representations. 
\paragraph{Instance code.}
Inspired by the works~\cite{jaegle2021perceiver, wang2021max}, which use a latent space to encode task-specific information, we introduce instance code $e$, a $L\times D$ vector to VIS task, where the $L$ is the maximum detected instance number in a frame and $D$ is the feature dimension for each instance.
Our instance code represents both the class and mask information of one instance for each slot in an order-aware fashion; thus, we can directly use slot indices to represent instance identities.
This differs our approaches from MaX-Deeplab~\cite{wang2021max} which takes the code as $L$ permutation invariant ones and~\cite{yang2019video} which only represents instance appearance information.
%Note that the $L$ slots in the instance code are order-aware, which is different from taking it as $L$ separated ones as MaX-Deeplab~\cite{wang2021max}. Instead of only representing instance appearance information \cite{yang2019video}, each slot directly represents an instance, which encodes both the class and mask information. And the slot index is defined to represent the instance identity.
\begin{figure}[h!]
\centering
\includegraphics[width=\linewidth]{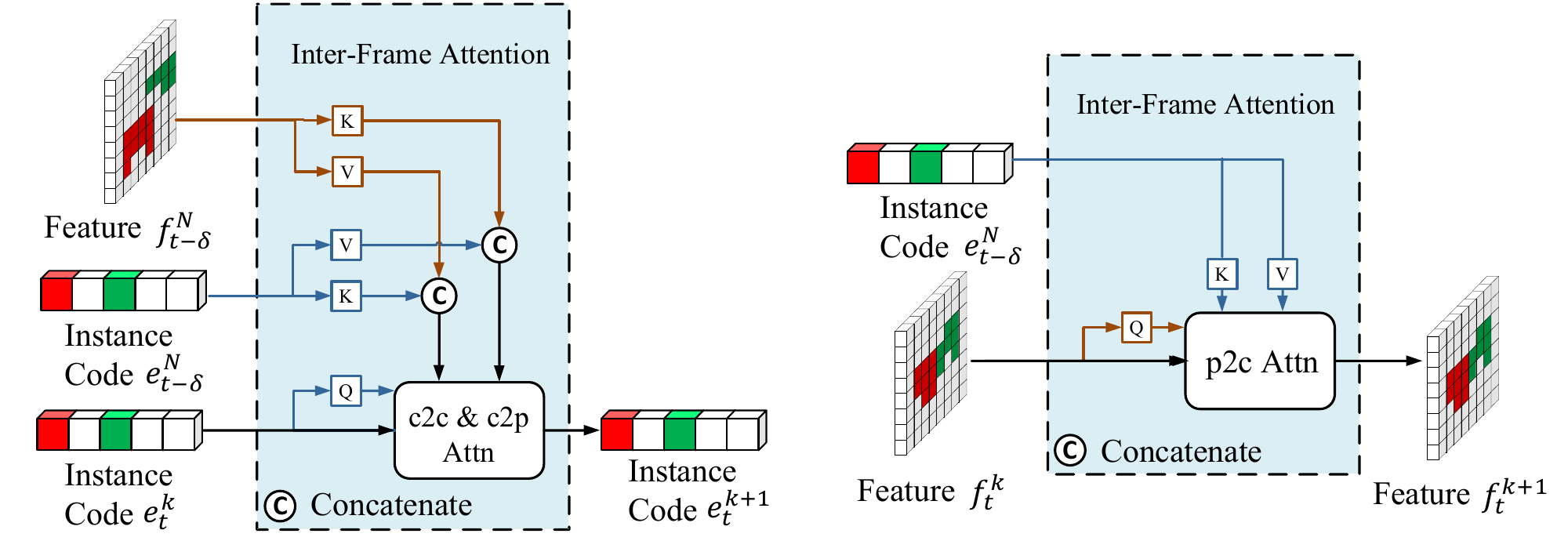}
\caption{\textbf{Inter-frame attention.} The intra-frame attention at  $k$-th layer enables communication between current frame features $e_t^k$, $f_t^k$ and past frame features $e_{t-\delta}^k$, $f_{t-\delta}^k$. The c2c\,\&\,c2p and p2c attentions fuse instance relevant features to instance code and pixel feature map respectively.}
\label{fig:inter_attn}
\end{figure}

\subsection{Instance-aware Frame Communication}
We enhance the frame representations with frame-wise correlation at both instance level and pixel level with multiple types of attention layers based on the transformer.
%To enable the instance-level fusion across frames, we bridge the reference and target feature maps with instance code to locate the same instance in the feature map throughout time. We construct the relation between instance code and feature map using intra-frame attention and fuse rich contextual information to both instance code and feature map by inter-frame attention.

\subsubsection{Intra-frame attention.}
Intra-frame attention layer constructs the relation between instance code and feature map by three types of attention - code-to-code (c2c), code-to-pixel (c2p), and pixel-to-code (p2c). The c2c and c2p attention aim to probe the instance relevant feature from pixel feature map and instance code respectively. The p2c attention adjusts the pixel feature map based on the instance code.

As shown in Figure~\ref{fig:intra_attn}, given a pixel feature map at $k$-th layer $f^k_t\in \mathbb{R}^{H\times W\times D_f}$ with height $H$, width $W$, dimension $D_f$ and an instance code $e^k_t\in \mathbb{R}^{L\times D_e}$ with length $L$, dimension $D_e$, we conduct the c2c and c2p attention simultaneously. We learn the query $Q_t^e$, key $K_t^e, K_t^f$ and value $V_t^e, V_t^f$ by linear projections from code $e^k_t$ and feature $f^k_t$ respectively. The output of intra-frame c2c \& c2p attention can be computed as
\begin{equation}
    e_{intra} = \sum^{HW+L}_{n=1} \mathrm{softmax}(Q_t^e \cdot K_t^{e\oplus f})V_t^{e\oplus f}
\label{equ:intra_prob}
\end{equation}
where $K_t^{e\oplus f}=[K_t^e; K_t^f]$ and $V_t^{e\oplus f}=[V_t^e; V_t^f]$. 
For simplicity, we will omit the layer numbers in Equation~\ref{equ:intra_prob} hereafter. Specifically, the output $e_{intra}$ is fused to input $e_t$ by residual connection.

Similarly, the output of intra-frame p2c attention can be computed as 
\begin{equation}
    f_{intra} = \sum^{L}_{n=1} \mathrm{softmax}(Q_t^f \cdot K_t^{e})V_t^{e} 
\end{equation}
Like intra-frame c2c \& c2p attention, the output of p2c attention $f_{intra}$ is fused to the input $f_t$ by residual connection.

\subsubsection{Inter-frame attention.}
Inter-frame attention aims to construct temporal correlation and fuse contextual features across frames.
Our key insight for inter-frame attention is that, given a clip of video frames, the same instance may appear in slightly different spatial localization while the appearance of it should be similar. Therefore, the instance code of adjacent frames should encode similar representations. To the end, we propose inter-frame c2c, c2p and p2c attentions to combine temporal information.

Specifically, given a target frame $I_t$ and a set of reference $\mathcal{R} = \{I_{t-\delta}\}$, $\delta = 1,\dots,N_{ref}$, we denote the instance code and feature map of target frame as $e_t$ and $f_t$; for reference frame, the instance code and feature map after the first $N$ intra-frame attention layers can be denote as $\{e^N_{t-\delta}\}$ and $\{f^N_{t-\delta}\}$, $\delta = 1,\dots,N_{ref}$.

%the extracted instance code and feature map of target frame can be denote as $e_t$ and $f_t$ respectively. Similarly, for reference frames, the instance code and feature map after the first $N$ intra-frame attention layers can be denote as $\{e^N_{t-\delta}\}$ and $\{f^N_{t-\delta}\}$, $\delta = 1,\dots,N_{ref}$.
To compute inter-frame attention with multiple reference frames, we first concatenate all $N_{ref}$ reference codes ($\{e^N_{t-\delta}\}_{\delta=1}^{N_{ref}}$) and feature maps ($\{f^N_{t-\delta}\}_{\delta=1}^{N_{ref}}$) and then add learnable positional encoding~\cite{dosovitskiy2020image} to them. Let us denote the processed reference inputs as $e^N_{ref}\in \mathbb{R}^{N_{ref}L\times D_e}$ and $f^N_{ref}\in \mathbb{R}^{N_{ref}H\times W\times D_f}$. 
Similarly, we add learnable positional encoding to the target inputs and denote them as $e_{tgt}\in \mathbb{R}^{L\times D_e}$ and $f_{tgt}\in \mathbb{R}^{H\times W\times D_f}$.
After that, we project the instance codes and feature maps to corresponding queries, keys, and values by linear transformations. Figure~\ref{fig:inter_attn} illustrates an example with $N_{ref}=1$ reference frame.

We fuse temporal information to target frame code by inter-frame c2c\,\&\,c2p attention. Specifically, a query $Q^e_{tgt}$ computed from target code is leveraged to probe the reference code and feature map. Let us denote keys and values from reference code and feature map as $K^e_{ref}$, $K^f_{ref}$ and $V^e_{ref}$, $V^f_{ref}$ respectively. The output of inter-frame c2c\,\&\,c2p attention can be computed as 
\begin{equation}
    e_{inter} = \sum^{HW+L}_{n=1} \mathrm{softmax}(Q_{tgt}^e \cdot K_{ref}^{e\oplus f})V_{ref}^{e\oplus f} 
\end{equation}
where $K_{ref}^{e\oplus f}=[K_{ref}^e; K_{ref}^f]$ and $V_{ref}^{e\oplus f}=[V_{ref}^e; V_{ref}^f]$.
To adjust pixel feature map accordingly, inter-frame p2c attention is utilized and can be computed as 
\begin{equation}
    f_{inter} = \sum^{L}_{n=1} \mathrm{softmax}(Q_{tgt}^f \cdot K_{ref}^{e})V_{ref}^{e} 
\end{equation}

Compared to previous pixel-level temporal attention modules \cite{fu2020compfeat, yang2020collaborative}, our code-based attention introduces the instance-aware fusion without redundant feature cropping, which enables us to leverage the feature-level consistency throughout time.

%(2) reduces the computational complexity from $\mathcal{O}(N_{ref}H^2W^2)$ to $\mathcal{O}(N_{ref}HWN_e)$ benefiting from instance code brigding. We will further discuss the effectiveness of inter-frame attention in Section~\ref{sec:inter-frame attn}.

\begin{table*}[htbp]
\centering
\scalebox{0.9}{
    \begin{tabular}{l|c|c|ccccc} 
    \hline
    \hline
    Method & Backbone & AP & AP50 & AP75 & AR1 & AR10\\
    \hline
    \multicolumn{7}{c}{Offline Methods} \\
    \hline
    STEm-Seg \cite{athar2020stem} & ResNet-50 & 30.6 & 50.7 & 33.5 & 31.6 & 37.1\\
    STEm-Seg \cite{athar2020stem} & ResNet-101 & 34.6 & 55.8 & 37.9 & 34.4 & 41.6\\
    MaskProp \cite{bertasius2020classifying} & ResNet-50 & 40.0 & - & 42.9 & - & - \\
    VisTR \cite{wang2021end} & ResNet-50$^\dagger$ & 36.2 & 59.8 & 36.9 & 37.2 & 42.4\\
    VisTR \cite{wang2021end} & ResNet-101$^\dagger$ & 40.1 & 64.0 & 45.0 & 38.3 & 44.9\\
    Purpose-Reduce \cite{lin2021video} & ResNet-50 & 40.4 & 63.0 & 43.8 & 41.1 & 49.7\\
    
    \hline
    \multicolumn{7}{c}{Online Methods} \\
    \hline
    MaskTrack R-CNN \cite{yang2019video} & ResNet-50 & 30.3 & 51.1 & 32.6 & 31.0 & 35.5\\
    MaskTrack R-CNN \cite{yang2019video} & ResNet-101 & 31.9 & 53.7 & 32.3 & 32.5 & 37.7\\
    SipMask \cite{cao2020sipmask} & ResNet-50 & 33.7 & 54.1 & 35.8 & 35.4 & 40.1\\
    CompFeat \cite{fu2020compfeat} & ResNet-50 & 35.3 & 56.0 & 38.6 & 33.1 & 40.3\\
    STMask \cite{li2021spatial} & ResNet-50 & 33.5 & 52.1 & 36.9 & 31.1 & 39.2\\
    STMask \cite{li2021spatial} & ResNet-101 &  36.8 & 56.8 & 38.0 & 34.8 & 41.8\\
    SG-Net \cite{liu2021sg} & ResNet-50 & 34.8 & 56.1 & 36.8 & 35.8 & 40.8\\
    SG-Net \cite{liu2021sg} & ResNet-101 & 36.3 & 57.1 & 39.6 & 35.9 & 43.0\\
    QueryInst \cite{fang2021instances} & ResNet-50 & 36.2 & 57.3 & 39.7 & 36.0 & 42.0\\
    CrossVIS \cite{yang2021crossover} & ResNet-50 & 36.3 & 56.8 & 38.9 & 35.6 & 40.7\\
    CrossVIS \cite{yang2021crossover} & ResNet-101 & 36.6 & 57.3 & 39.7 & 36.0 & 42.0\\
    \hline
    \textbf{Ours} & ResNet-50$^\ddagger$ & 41.3 &61.5 & 43.5 & 42.7 & 47.8\\
    \hline
    \hline
    \end{tabular}
}
\caption{Comparison to state-of-the-art video instance segmentation on \textbf{Youtube-VIS-2019} val set. ResNet-50$^\dagger$ and ResNet-101$^\dagger$ means transformer on top of ResNet-50 and ResNet-101 respectively. ResNet-50$^\ddagger$ indicates ResNet-50 with intra-frame and inter-frame layers.
}
\label{tab:ytvis2019}
\vspace{-0.3cm}
\end{table*}

\subsection{Network Design}
Our method employs an encoder-decoder-based structure equipped with a transformer integrated in the encoder.

\subsubsection{Encoder.}
Unlike previous transformer-based VIS methods that overlap a transformer on top of the backbone, we insert the intra-frame attention layers into the last stage of the ResNet-50~\cite{he2016deep} backbone to better extract instance code. Following \cite{wang2021max}, we also use additional pixel-wise attention in intra-frame layers in the backbone to better extract global information for segregated or partially occluded masks in the VIS task.

After intra-frame attention, both the instance code and pixel-level feature maps are fed into the inter-frame attention layer to fuse temporal context. Following the conventional transformer, we iteratively repeat the inter-frame and intra-frame attention $M$ times before the decoder.

\subsubsection{Decoder.}
We follow the decoder structure of Deeplab-V3+~\cite{chen2018encoder} to fuse the low-level features. From the decoder features, we predict instance classes and instance masks separately. For instance class prediction, two fully connected layers and a softmax are applied on instance code before the final instance class output. For instance mask prediction, we leverage dynamic convolution ~\cite{jia2016dynamic} to correspond the instance code with masks. 
In particular, we learn dynamic filters $\theta_t\in \mathbb{R}^{N_e\times D_{fout}}$ from instance code by another two fully connection layers, where $N_e$ is the length of instance code and $D_{fout}$ the dimension of the upsampled feature map. Let the upsampled feature map be $f_{out}\in \mathbb{R}^{H_o\times W_o\times D_{fout}}$. The mask prediction $M_t\in \mathbb{R}^{N_e\times H_o\times W_o}$ can be compute as 
\begin{equation}
    M_t = softmax(\theta_tf_{out}^T) 
\end{equation}
where the softmax is applied on the $N_e$ dimension.

\subsection{Loss Function}
As an online VIS method, one main challenge of our method is to maintain the instance identity consistent throughout the video sequence. 
%Since IATF is an online VIS method, one of the main challenges is maintaining the instance identity consistent throughout the video sequence.
To tackle this problem, we consider the instance predictions in both current and past frames during training.
%when giving the network supervision.

To train the network, we assign a ground-truth to each instance prediction by searching a permutation of $L$ elements $\sigma\in\mathcal{S}_L$ with the highest similarity. Let us denote the $L$ instance predictions in arbitrary time $\tau$ as $\{y^i_\tau\}_{i=1}^{L}=\{p^i_\tau(c), m^i_\tau\}_{i=1}^{L}$ and the ground-truths $\{\hat{y}_\tau\}_{i=1}^{L}=\{\hat{c}^i_\tau, \hat{m}^i_\tau\}_{i=1}^{L}$ , where $p_\tau^i(c)$ and $m_\tau^i$ represents the probability of class $c$ (including $\emptyset$) and mask of the $i$-th instance respectively. Since VIS tasks assume the prediction as true positive only if it has accurate class prediction as well as mask prediction, we calculate the similarity as 
\begin{equation}
    \mathrm{Sim} = \mathrm{Sim}_{mask}\times \mathrm{Sim}_{cls}
\end{equation}
where the $\mathrm{Sim}_{mask}$ and $\mathrm{Sim}_{cls}$ are similarity in terms of mask and class respectively. To retain the order of instance code, we consider $t$ and $t-1$ frames simultaneously when compute the similarity. In particular, the $\mathrm{Sim}_{mask}$ and $\mathrm{Sim}_{cls}$ can be computed as

\begin{equation}\mathrm{Sim}_{mask}=\mathbbm{1}_{\{c^i_{t}\neq\emptyset\}}\mathrm{Dice}([m^{\sigma(i)}_t, m^{\sigma(i)}_{t-1}], [\hat{m}^i_t, \hat{m}^i_{t-1}])
\end{equation}

\begin{equation}
\mathrm{Sim}_{cls}=\mathbbm{1}_{\{c^i_{t}\neq\emptyset\}}[p_t^{\sigma(i)}(\hat{c}_t^i)+p_{t-1}^{\sigma(i)}(\hat{c}_{t-1}^i)]
\end{equation}
where $\mathbbm{1}_{\{\cdot\}}$ is an indicator function and $\sigma$ is a permutation of slot indices. $\mathrm{Dice}$ indicates the Dice loss \cite{milletari2016v}.

Given the optimal assignment $\hat{\sigma}$, the total loss $\mathcal{L}$ can be boiled down to three main components, $\mathcal{L}_{pos}$ for the $K$ matched instance predictions, $\mathcal{L}_{neg}$ for the $L-K$ unmatched instance predictions, and auxiliary losses $\mathcal{L}_{aux}$ \cite{wang2021max} to facilitate the training:
\begin{equation}
    \mathcal{L} = \lambda_{inst}\mathcal{L}_{pos} + (1-\lambda_{inst})\mathcal{L}_{neg} + \lambda_{aux}\mathcal{L}_{aux}
\end{equation}
where $\lambda_{inst}$ and $\lambda_{aux}$ are scalars to balance the losses, $\mathcal{L}_{neg}=-\sum_{K+1}^{N_e}\log p_t^{\hat{\sigma}(i)}(\emptyset)$ and $\mathcal{L}_{pos} = -\sum_{i=1}^K[k_{mask}\cdot\mathrm{Dice(m_t^{\hat{\sigma}(i)}, \hat{m}_t^{i})}+k_{cls}\cdot\log p_t^{\hat{\sigma}(i)}(\hat{c}_i)]$. The $k_{mask}$ and $k_{cls}$ are weights for mask term and class term respectively.

\subsection{Instance Identity Matching}
Different from previous online methods that crop instances then leverage multiple cues to match instances across frames, we directly assume the non-empty predictions from the same slot of the instance code have the same identity. To enhance model robustness, if the mask in $i$-th slot in time $t$ has an IoU larger than 0.5 with mask in $j$-th slot in time $t-1$, we directly assume they share the same identity without considering the slot indices.

\begin{figure*}[h!]
\centering
\includegraphics[width=0.9\linewidth, height=0.6\linewidth]{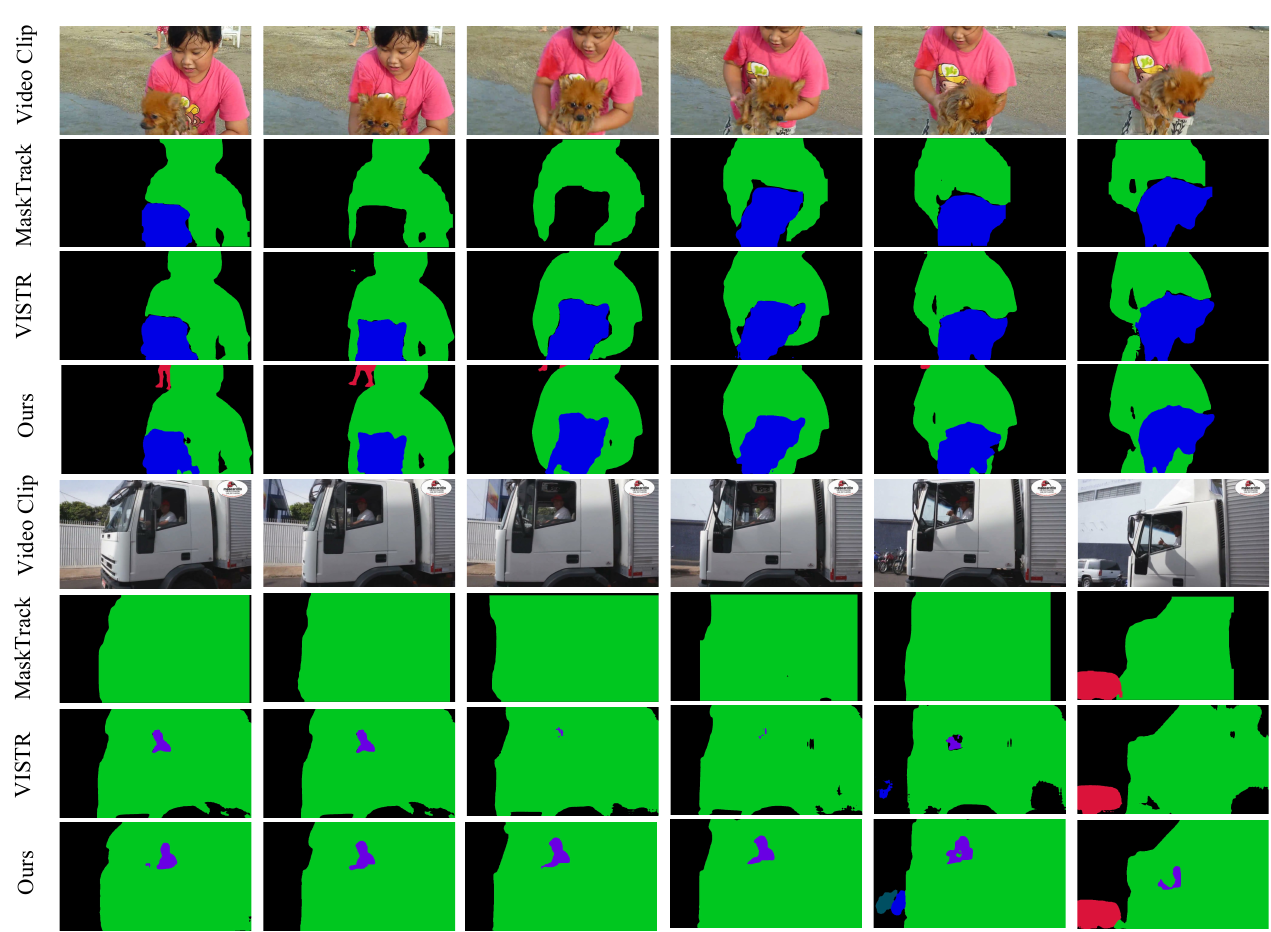}
\caption{Comparison to state-of-the-art video instance segmentation methods, MaskTrack \cite{yang2019video} and VISTR \cite{wang2021end}. Each row represents the same video clip. For each clip, colors indicate instance identities (best viewed in color). We show that our method generates more accurate and temporally consistent results, while MaskTrack R-CNN and VISTR attempt to miss instances for the cases where instances are overlapped or small.}
\label{fig:vis_seg}
\vspace{-0.1cm}
\end{figure*}

\section{Experiment}
\subsection{Dataset and Evaluation Metric}
We evaluate our method on two extensively used VIS datasets - Youtube-VIS-2019 and Youtube-VIS-2021.
\begin{itemize}
    \item \textbf{Youtube-VIS-2019} Youtube-VIS-2019 has 40 categories, 4,883 unique video instances, and 131k high-quality manual annotations. There are 2,238 training videos, 302 validation videos, and 343 test videos in it.
    \item \textbf{Youtube-VIS-2021} Youtube-VIS-2021 is an improved version of the Youtube-VIS-2019 dataset, which contains 8,171 unique video instances and 232k high-quality manual annotations. There are 2,985 training videos, 421 validation videos, and 453 test videos in this dataset.
\end{itemize}
The evaluation metric for this task is defined as the area under the precision-recall curve with different IoUs as thresholds.

\begin{table*}[h]
%%%%%%%%%%%%%%%%%%%%%%%%%%%%%%%%%%%%%
\begin{minipage}[h]{\textwidth}
%%%%%%%%%%%%%%%%%%%%%%%%%%%%%%%%%%%%%
\begin{minipage}[t]{0.35\textwidth}
\makeatletter\def\@captype{table}
\centering
\scalebox{0.7}{
    \begin{tabular}{cccccc}
    \hline
    \hline
    Method & AP & AP\@50 & AP\@75 & AR\@1 & AR\@10 \\
    \hline
    MaskTrack R-CNN & 28.6 & 48.9 & 29.6 & 26.5 & 33.8\\
    SipMask-VIS & 31.7 & 52.5 & 34.0 & 30.8 & 37.8\\
    CrossVIS & 34.2 & 54.4 & 37.9 & 30.4 & 38.2\\
    Ours & 35.8 & 56.3 & 39.1 & 33.6 & 40.3\\
    \hline
    \hline
    \end{tabular}}
    \caption{Comparison to state-of-the-art video instance segmentation on \textbf{Youtube-VIS-2021} val set. All results presented are generated by ResNet-50 backbone.}
    \label{tab:ytvis2021}
\end{minipage}
\hspace{0.02\textwidth}
\begin{minipage}[t]{0.3\textwidth}
\makeatletter\def\@captype{table}
\centering
\scalebox{0.7}{
    \begin{tabular}{ccccccc} 
    \hline
    \hline
    PM & p2c & c2c\&c2p & AP & AP\@50 & AP\@75\\
    \hline
    \Checkmark & & & 36.9 & 57.9 & 40.5\\
    & \Checkmark & & 38.8 & 59.7 & 40.4 \\
    \Checkmark & \Checkmark & & 39.4 & 60.6 & 41.4\\
    &\Checkmark & \Checkmark & 40.7 & 61.3 & 42.7\\
    \Checkmark &\Checkmark & \Checkmark & 41.3 & 61.5 & 43.5\\
    \hline
    \hline
    \end{tabular}}
    \caption{\textbf{Inter-frame attention.} PM, p2c and c2c\&c2p represents pair-wise matching strategy, inter-frame p2c attention and inter-frame c2c \& c2p attention respectively. \Checkmark indicates enabling the component.} 
    \label{tab:inter_attn}
\end{minipage}
\hspace{0.02\textwidth}
\begin{minipage}[t]{0.33\textwidth}
\makeatletter\def\@captype{table}
\centering
\scalebox{0.7}{
    \begin{tabular}{cccccc}
    \hline
    \hline
    Ref. Number & AP & AP\@50 & AP\@75 & AR\@1 & AR\@10 \\
    \hline
    1 & 39.7 & 60.9 & 41.2 & 39.8 & 44.8\\
    2 & 40.5 & 59.7 & 42.1 & 42.4 & 46.7\\
    3 & 41.3 & 61.5 & 43.5 & 42.7 & 47.8\\
    4 & 41.3 & 61.7 & 43.5 & 42.0 & 46.7\\
    \hline
    \hline
    \end{tabular}}
    \caption{\textbf{Reference frame number.} The reference frames are selected from the adjacent frames before target frame.}
    \label{tab:ref_num}
\end{minipage}

\end{minipage} 
\end{table*}

\begin{table*}[ht]
%%%%%%%%%%%%%%%%%%%%%%%%%%%%%%%%%%%%%
\begin{minipage}[t]{\textwidth}
\centering
%%%%%%%%%%%%%%%%%%%%%%%%%%%%%%%%%%%%%
\begin{minipage}[h]{0.47\textwidth}
\makeatletter\def\@captype{table}
\centering
\scalebox{1.0}{
    \begin{tabular}{cccccc}
    \hline
    \hline
    Layer Number & AP & AP\@50 & AP\@75 & AR\@1 & AR\@10 \\
    \hline
    1 & 40.1 & 61.1 & 41.9 & 41.4 & 46.2\\
    2 & 41.3 & 61.5 & 43.5 & 42.7 & 47.8\\
    3 & 40.8 & 61.8 & 42.0 & 41.9 & 46.4\\
    4 & 40.2 & 61.0 & 41.9 & 41.6 & 46.1\\
    \hline
    \hline
    \end{tabular}}
    \caption{\textbf{Alternate attention layers.} Note that the first $N$ intra-frame attention are not included in the layer number.}
    \label{tab:TrLayerNum}
\end{minipage}
\hspace{0.04\textwidth}
\begin{minipage}[h]{0.47\textwidth}
\makeatletter\def\@captype{table}
\centering
\scalebox{1.0}{
    \begin{tabular}{cccccc}
    \hline
    \hline
    Slot Number & AP & AP\@50 & AP\@75 & AR\@1 & AR\@10 \\
    \hline
    10 & 41.3 & 61.5 & 43.5 & 42.7 & 47.8\\
    15 & 41.2 & 61.9 & 43.5 & 42.3 & 47.6\\
    20 & 41.1 & 62.6 & 42.5 & 41.9 & 47.8\\
    25 & 41.3 & 61.2 & 44.0 & 42.7 & 47.2\\
    \hline
    \hline
    \end{tabular}}
    \caption{\textbf{Slot number of instance embedding.} Youtube-VIS-2019 has a maximum of 10 and a minimum of 1 instance in a frame.}
    \label{tab:slotNum}
\end{minipage}
\end{minipage}
\vspace{-0.1cm}
\end{table*}

\subsection{Implementation Details}
\label{sec:impelementation}
We implement our method with the Tensorflow2 framework. Following previous methods \cite{fu2020compfeat, lin2021video}, we first pre-train our model with both Youtube-VIS and overlapped categories on COCO dataset \cite{lin2014microsoft} then finetune the model on the Youtube-VIS dataset. All frames are resized and padded to $641\times 641$ during training and inference. We train our model for 35k iterations with a “poly” learning rate policy where the learning rate is multiplied by $(1-\frac{iter}{iter_{max}})^{0.9}$ for each iteration with an initial learning rate of 0.001 to all experiments. The $\mathrm{batchsize}=32$ and an adam \cite{kingma2014adam} optimizer with $\beta_1=0.9$, $\beta_2=0.999$ and $\mathrm{weight\, decay}=0$ is leveraged. Multi-scale training is adopted to obtain a strong baseline. We select adjacent three frames as reference frames if not specific.

\subsection{Main Results}
In this section, we compare our method with previous state-of-art methods in terms of Youtube-VIS 2019 and Youtube-VIS 2021 datasets.

\subsubsection{Quantitative results.}
We compare our method against state-of-the-art VIS methods on Youtube-VIS 2019 dataset in Table~\ref{tab:ytvis2019}. (1) Compared with online methods. Notably, our method achieves a mAP of 41.3 which is the best among all online methods and eclipses the previous state-of-art online method CrossVIS by a large margin even if it is with a stronger ResNet-101 backbone. (2) Compared with offline methods. Our method also outperforms all offline VIS methods when using the same ResNet-50 backbone. For another transformer-based method VisTR, which leverages a transformer on top of the backbone and matches instance with a sequential level strategy, our method outperforms it by 4.3 mAP when using the same ResNet-50 backbone and 1.2 mAP when it is using the stronger ResNet-101 backbone. Moreover, we also report the results on the recently introduced Youtube-VIS 2021 dataset in Table~\ref{tab:ytvis2021}. Our method also achieves the best result, 35.8 mAP, on the newly imported dataset. 

\subsubsection{Qualitative results.}
We present our qualitative result in Figure~\ref{fig:vis_seg} and compare it against VisTR \cite{wang2021end}. The result indicates that VisTR fails to segment and track instances when they are overlapped. In contrast, our method shows great accuracy and robustness even in very challenging scenarios. This implies that our transformer-based network equipped with instance code generates more accurate and temporally consistent results than simply adopting conventional transformer architecture \cite{vaswani2017attention} on top of the conventional backbone.

\subsection{Ablation Study}
We conduct extensive ablation studies on Youtube-VIS-2019 to show the effectiveness of different components of our method. 

\subsubsection{Inter-frame attentions.}
\label{sec:inter-frame attn}
To investigate the effectiveness of the inter-frame attention layer, we train models by disabling different attentions. As shown in Table~\ref{tab:inter_attn}, the model performance plummets to 36.9 mAP when disabling all inter-frame attentions. If only adopting inter-frame p2c attention, the performance will also drop about 2.0 mAP compared to the default setting. In addition, we found that when we leverage all attention types in the inter-frame layer, the pair-wise matching strategy only brings a marginal gain to the overall performance. This indicates that by using the inter-frame attentions, the model can automatically learn temporal consistency without additional supervision.

\subsubsection{Reference frame number.}
To investigate the importance of the temporal information to our method, we conduct experiments by teasing the reference frame number to the inter-frame attention layer. As illustrated in Table~\ref{tab:ref_num}, with the reference frame number varying from 1 to 4, the mAP of our method gradually increases from 39.7 to 41.3, which verifies the necessity of the temporal information to our model. 

\subsubsection{Alternate temporal transformer layer number.}
We design our temporal transformer referring to the conventional transformer architecture \cite{vaswani2017attention}, where the inter-frame layers in reference frames correspond to the conventional transformer encoder and the alternate transformer layers in the target frame correspond to the conventional transformer decoder. We ablate on the alternate transformer layers number to investigate how many layers are enough to fuse the temporal features. Table~\ref{tab:TrLayerNum} shows the results of different numbers of transformer layers. We found that a small number of 2 leads to the best performance. This demonstrates that it is easier to learn better features by using a swallow temporal fusion network.

\subsubsection{Mask slot number.}
The slot number of instance code represents the maximum number of detected instances in a frame. When the instance number in a frame is smaller than the slot number, the remaining unmatched slots will predict empty class and be assumed as negative samples to be supervised separately. Since the inter-frame attention calculates the attention map separately for each slot in the instance code while conducting softmax among all slots, the redundant slots in the instance code may influence the overall performance. However, as shown in Table~\ref{tab:slotNum}, we found the performance retaining robust even with numerous redundant slots existing. The results indicate that both intra-frame and inter-frame attentions are robust to redundant slots. This property enables our method to handle the scenario that has extremely few instances.

\section{Conclusion}
In this paper, we propose a novel instance-aware temporal fusion method for VIS, which enables the temporal fusion at hybrid level (pixel-, instance-, cross-level). Based on that, we further simplify the instance identity association operation. Notably, our method achieves the best result among both Youtube-VIS-2019 and Youtube-VIS-2021 benchmarks. Moreover, extensive study shows that our instance-aware temporal fusion leads to remarkable improvement to the VIS performance.

\newpage

{\small
\bibliography{aaai22.bib}
}
\end{document}